%% file: main.tex
\documentclass[10pt,twocolumn,letterpaper]{article}

\usepackage{iccv}
\usepackage{times}
\usepackage{epsfig}
\usepackage{graphicx}
\usepackage{amsmath}
\usepackage{amssymb}

\usepackage{adjustbox}
\usepackage{array}
\usepackage{booktabs}
\usepackage{colortbl}
\usepackage{float,wrapfig}
\usepackage{hhline}
\usepackage{multirow}
\usepackage{subcaption} 
\usepackage{enumitem}
\usepackage[accsupp]{axessibility}  

\usepackage[pagebackref=true,breaklinks=true,colorlinks,bookmarks=false]{hyperref}

\iccvfinalcopy 

\def\iccvPaperID{12517} 

\ificcvfinal\fi

\def\modelshort{EfficientViT\xspace}
\def\modela{\modelshort-B0\xspace}
\def\modelb{\modelshort-B1\xspace}
\def\modelc{\modelshort-B2\xspace}
\def\modeld{\modelshort-B3\xspace}

\DeclareMathOperator*{\mr}{\mathbb{R}}

\begin{document}

\title{EfficientViT: Multi-Scale Linear Attention for High-Resolution Dense Prediction}

\author{
    Han Cai$^1$, Junyan Li$^2$, Muyan Hu$^3$, Chuang Gan$^4$, Song Han$^1$ \\
    $^1$MIT, $^2$Zhejiang University, $^3$Tsinghua University, $^4$MIT-IBM Watson AI Lab \\
    \url{https://github.com/mit-han-lab/efficientvit}
}

\maketitle
\ificcvfinal\thispagestyle{empty}\fi

\begin{abstract}
High-resolution dense prediction enables many appealing real-world applications, such as computational photography, autonomous driving, etc. However, the vast computational cost makes deploying state-of-the-art high-resolution dense prediction models on hardware devices difficult. This work presents EfficientViT, a new family of high-resolution vision models with novel multi-scale linear attention. Unlike prior high-resolution dense prediction models that rely on heavy softmax attention, hardware-inefficient large-kernel convolution, or complicated topology structure to obtain good performances, our multi-scale linear attention achieves the global receptive field and multi-scale learning (two desirable features for high-resolution dense prediction) with only lightweight and hardware-efficient operations. As such, EfficientViT delivers remarkable performance gains over previous state-of-the-art models with significant speedup on diverse hardware platforms, including mobile CPU, edge GPU, and cloud GPU. 
Without performance loss on Cityscapes, our EfficientViT provides up to 13.9$\times$ and 6.2$\times$ GPU latency reduction over SegFormer and SegNeXt, respectively. For super-resolution, EfficientViT delivers up to 6.4$\times$ speedup over Restormer while providing 0.11dB gain in PSNR. For Segment Anything, EfficientViT delivers 48.9$\times$ higher throughput on A100 GPU while achieving slightly better zero-shot instance segmentation performance on COCO. 
\end{abstract}

\input{sections/1-intro}

\input{sections/3-method}

\input{sections/4-exp}

\input{sections/2-related-work}

\input{sections/5-conclusion}

{\small
\bibliographystyle{unsrt}
\bibliography{egbib}
}

\end{document}

%% file: sections/1-intro.tex
\section{Introduction}
High-resolution dense prediction is a fundamental task in computer vision and has broad applications in the real world, including autonomous driving, medical image processing, computational photography, etc. Therefore, deploying state-of-the-art (SOTA) high-resolution dense prediction models on hardware devices can benefit many use cases. 

However, there is a large gap between the computational cost required by SOTA high-resolution dense prediction models and the limited resources of hardware devices. It makes using these models in real-world applications impractical. In particular, high-resolution dense prediction models require high-resolution images and strong context information extraction ability to work well \cite{badrinarayanan2017segnet,ronneberger2015u,chen2017deeplab,zhao2017pyramid,yuan2020object,wang2020deep}. Therefore, directly porting efficient model architectures from image classification is unsuitable for high-resolution dense prediction. 

\begin{figure*}[t]
    \centering
    \includegraphics[width=\linewidth]{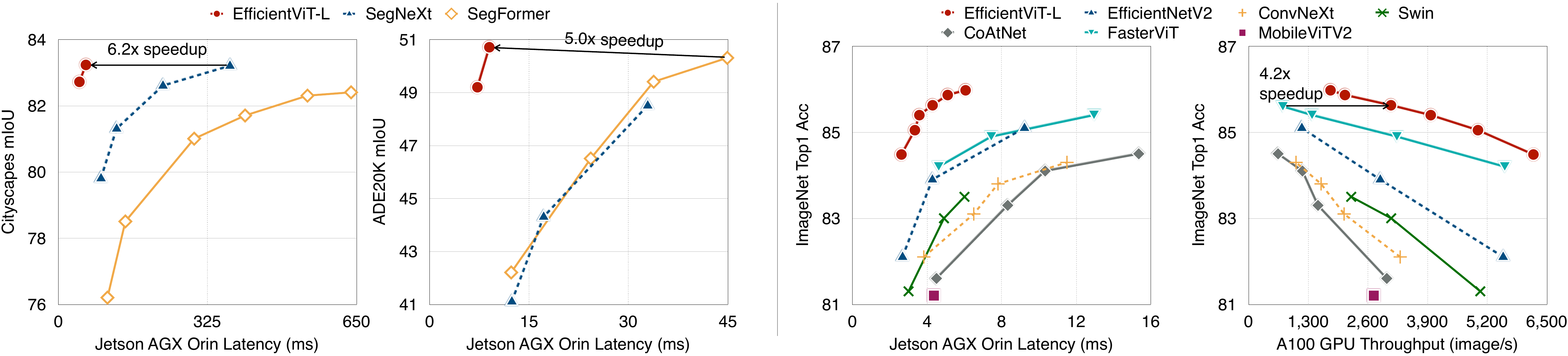}
    \caption{\textbf{Latency/Throughput vs. Performance.} All performance results are obtained with the single model and single-scale inference. The GPU latency/throughput results are obtained on one edge GPU (Jetson AGX Orin) and one cloud GPU (A100) using TensorRT and fp16. EfficientViT consistently achieves a remarkable boost in speed on diverse hardware platforms while providing the same/higher performances on Cityscapes, ADE20K, and ImageNet than prior segmentation/classification models.}
    \label{fig:figure1}
\end{figure*}

This work introduces \textbf{EfficientViT}, a new family of vision transformer models for efficient high-resolution dense prediction. The core of EfficientViT is a new multi-scale linear attention module that enables the global receptive field and multi-scale learning with hardware-efficient operations. 
Our module is motivated by prior SOTA high-resolution dense prediction models. They demonstrate that the multi-scale learning \cite{chen2017deeplab,zhao2017pyramid} and global receptive field \cite{xie2021segformer} are critical in improving models' performances. However, they do not consider hardware efficiency when designing their models, which is essential for real-world applications. 
For example, SegFormer \cite{xie2021segformer} introduces softmax attention \cite{vaswani2017attention} into the backbone to have a global receptive field. However, its computational complexity is quadratic to the input resolution, making it unable to handle high-resolution images efficiently. 
SegNeXt \cite{guo2022segnext} proposes a multi-branch module with large-kernel convolutions (kernel size up to 21) to enable a large receptive field and multi-scale learning. However, large-kernel convolution requires exceptional support on hardware to achieve good efficiency \cite{ding2022scaling,wang2022lite}, which is usually unavailable on hardware devices. 

Hence, the design principle of our module is to enable these two critical features while avoiding hardware-inefficient operations. Specifically, we propose substituting the inefficient softmax attention with lightweight ReLU linear attention \cite{katharopoulos2020transformers} to have the global receptive field. By leveraging the associative property of matrix multiplication, ReLU linear attention can reduce the computational complexity from quadratic to linear while preserving functionality. In addition, it avoids hardware-inefficient operations like softmax, making it more suitable for hardware deployment (Figure~\ref{fig:latency_linear_vs_softmax}). 

However, ReLU linear attention alone has limited capacity due to the lack of local information extraction and multi-scale learning ability. Therefore, we propose to enhance ReLU linear attention with convolution and introduce the multi-scale linear attention module to address the capacity limitation of ReLU linear attention. Specifically, we aggregate nearby tokens with small-kernel convolutions to generate multi-scale tokens. We perform ReLU linear attention on multi-scale tokens (Figure~\ref{fig:block}) to combine the global receptive field with multi-scale learning. We also insert depthwise convolutions into FFN layers to further improve the local feature extraction capacity. 

We extensively evaluate EfficientViT on two popular high-resolution dense prediction tasks: semantic segmentation and super-resolution. EfficientViT provides significant performance boosts over prior SOTA high-resolution dense prediction models. More importantly, EfficientViT does not involve hardware-inefficient operations, so our \#FLOPs reduction can easily translate to latency reduction on hardware devices (Figure~\ref{fig:figure1}). 

In addition to these conventional high-resolution dense prediction tasks, we apply EfficientViT to Segment Anything \cite{kirillov2023segment}, an emerging promptable segmentation task that allows zero-shot transfer to many vision tasks. EfficientViT achieves 48.9$\times$ acceleration on A100 GPU than SAM-ViT-Huge \cite{kirillov2023segment} without performance loss. 
We summarize our contributions as follows:
\begin{itemize}[leftmargin=*]
\item We introduce a new multi-scale linear attention module for efficient high-resolution dense prediction. It achieves the global receptive field and multi-scale learning while maintaining good efficiency on hardware. To the best of our knowledge, our work is the first to demonstrate the effectiveness of linear attention for high-resolution dense prediction.

\item We design EfficientViT, a new family of high-resolution vision models, based on the proposed multi-scale linear attention module. 

\item Our model demonstrates remarkable speedup on semantic segmentation, super-resolution, Segment Anything, and ImageNet classification on diverse hardware platforms (mobile CPU, edge GPU, and cloud GPU) over prior SOTA models.
\end{itemize}

%% file: sections/3-method.tex
\begin{figure*}[t]
    \centering
    \includegraphics[width=\linewidth]{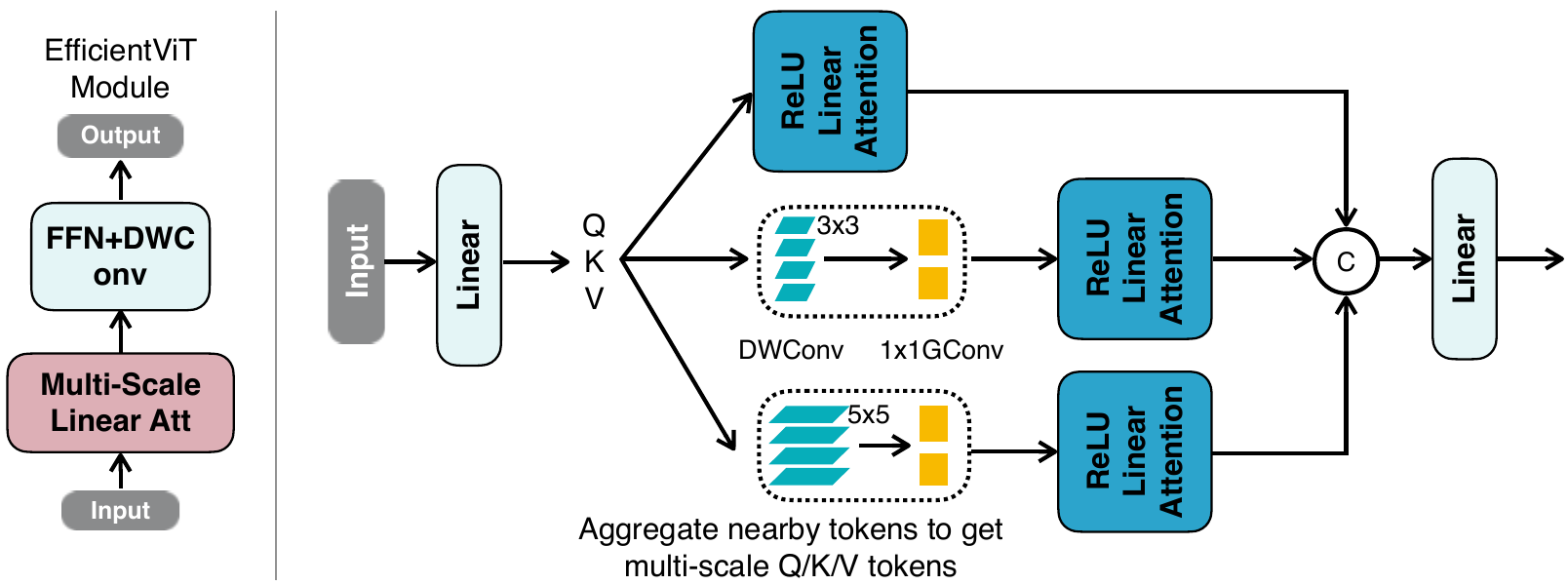}
    \caption{\textbf{EfficientViT's Building Block (left) and Multi-Scale Linear Attention (right).} \emph{Left}: EfficientViT's building block consists of a multi-scale linear attention module and an FFN with depthwise convolution (FFN+DWConv). Multi-scale linear attention is responsible for capturing context information, while FFN+DWConv captures local information. \emph{Right}: After getting Q/K/V tokens via the linear projection layer, we generate multi-scale tokens by aggregating nearby tokens via lightweight small-kernel convolutions. ReLU linear attention is applied to multi-scale tokens, and the outputs are concatenated and fed to the final linear projection layer for feature fusing. 
    }
    \label{fig:block}
\end{figure*}

\begin{figure*}[t]
    \centering
    \includegraphics[width=\linewidth]{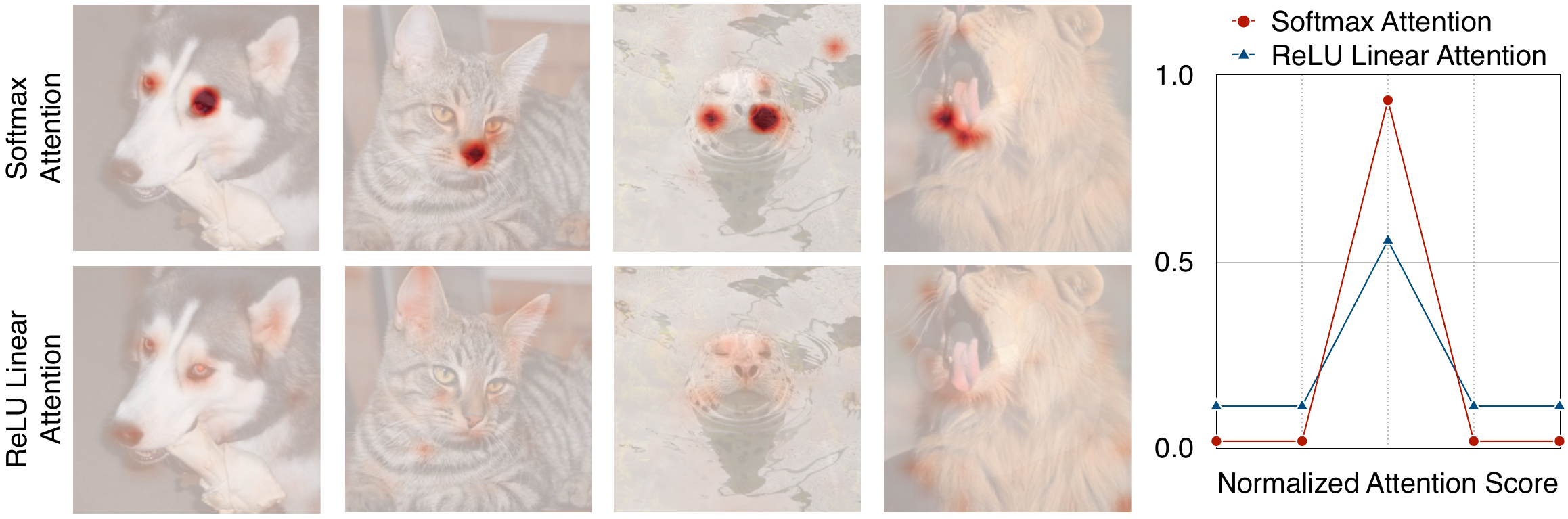}
    \caption{\textbf{Softmax Attention vs. ReLU Linear Attention.} Unlike softmax attention, ReLU linear attention cannot produce sharp attention distributions due to a lack of the non-linear similarity function. Thus, its local information extraction ability is weaker than the softmax attention. 
    }
    \label{fig:linear_limitation}
\end{figure*}

\begin{figure}[t]
    \centering
    \includegraphics[width=0.7\linewidth]{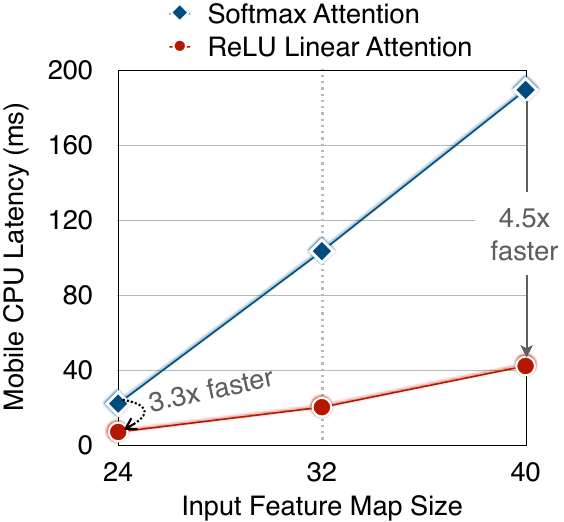}
    \caption{\textbf{Latency Comparison Between Softmax Attention and ReLU Linear Attention.} ReLU linear attention is 3.3-4.5$\times$ faster than softmax attention with similar computation, thanks to removing hardware-unfriendly operations (e.g., softmax). Latency is measured on the Qualcomm Snapdragon 855 CPU with TensorFlow-Lite, batch size 1, and fp32.}
    \label{fig:latency_linear_vs_softmax}
\end{figure}

\begin{figure*}[t]
    \centering
    \includegraphics[width=\linewidth]{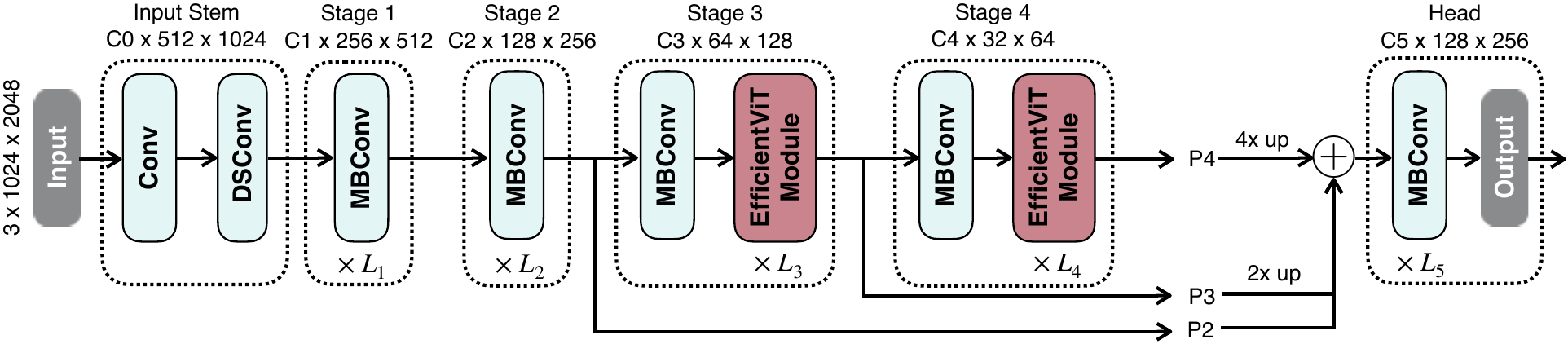}
    \caption{\textbf{Macro Architecture of EfficientViT.} We adopt the standard backbone-head/encoder-decoder design. We insert our EfficientViT modules in Stages 3 and 4 in the backbone. Following the common practice, we feed the features from the last three stages (P2, P3, and P4) to the head. We use addition to fuse these features for simplicity and efficiency. We adopt a simple head design that consists of several MBConv blocks and output layers.}
    \label{fig:macro_arch}
\end{figure*}

\section{Method}
This section first introduces the multi-scale linear attention module. Unlike prior works, our multi-scale linear attention simultaneously achieves the global receptive field and multi-scale learning with only hardware-efficient operations. Then, based on the multi-scale linear attention, we present a new family of vision transformer models named EfficientViT for high-resolution dense prediction.

\subsection{Multi-Scale Linear Attention} 
Our multi-scale linear attention balances two crucial aspects of efficient high-resolution dense prediction, i.e., performance and efficiency. Specifically, the global receptive field and multi-scale learning are essential from the performance perspective. Previous SOTA high-resolution dense prediction models provide strong performances by enabling these features but fail to provide good efficiency. Our module tackles this issue by trading slight capacity loss for significant efficiency improvements. 

An illustration of the proposed multi-scale linear attention module is provided in Figure~\ref{fig:block} (right). In particular, we propose to use ReLU linear attention \cite{katharopoulos2020transformers} to enable the global receptive field instead of the heavy softmax attention \cite{vaswani2017attention}. While ReLU linear attention \cite{katharopoulos2020transformers} and other linear attention modules \cite{bolya2023hydra,choromanski2020rethinking,shen2021efficient,wang2020linformer} have been explored in other domains, it has never been successfully applied to high-resolution dense prediction. To the best of our knowledge, EfficientViT is the first work demonstrating ReLU linear attention's effectiveness in high-resolution dense prediction. In addition, our work introduces novel designs to address its capacity limitation. 

\vspace{-10pt}
\paragraph{Enable Global Receptive Field with ReLU Linear Attention.} 

Given input $x \in \mr^{N \times f}$, the generalized form of softmax attention can be written as: 
\begin{equation}\label{eq:generalizedSoftmaxAtt}\small
    O_i = \sum_{j=1}^N\frac{Sim(Q_i, K_j)}{\sum_{j=1}^N{Sim(Q_i, K_j)}}V_j, 
\end{equation}
where $Q=xW_Q$, $K=xW_K$, $V=xW_V$ and $W_Q/W_K/W_V \in \mr^{f \times d}$ is the learnable linear projection matrix. $O_i$ represents the i-th row of matrix $O$. $Sim(\cdot , \cdot)$ is the similarity function. When using the similarity function $Sim(Q, K) = \exp(\frac{Q K^T}{\sqrt{d}})$, Eq.~(\ref{eq:generalizedSoftmaxAtt}) becomes the original softmax attention \cite{vaswani2017attention}. 

Apart from $\exp(\frac{Q K^T}{\sqrt{d}})$, we can use other similarity functions. In this work, we use ReLU linear attention \cite{katharopoulos2020transformers} to achieve both the global receptive field and linear computational complexity. In ReLU linear attention, the similarity function is defined as
\begin{equation}\small
    Sim(Q, K) = \text{ReLU}(Q) \text{ReLU}(K)^T.
\end{equation}

With $Sim(Q, K) = \text{ReLU}(Q) \text{ReLU}(K)^T$, Eq.~(\ref{eq:generalizedSoftmaxAtt}) can be rewritten as:
{\small
\begin{align*}
    O_i &= \sum_{j=1}^N\frac{\text{ReLU}(Q_i) \text{ReLU}(K_j)^T}{\sum_{j=1}^N{{\color{red}\text{ReLU}(Q_i)} \text{ReLU}(K_j)^T}}V_j \\ 
    &= \frac{\sum_{j=1}^N (\text{ReLU}(Q_i) \text{ReLU}(K_j)^T) V_j}{{\color{red}\text{ReLU}(Q_i)} \sum_{j=1}^N \text{ReLU}(K_j)^T}.
\end{align*}
}

Then, we can leverage the associative property of matrix multiplication to reduce the computational complexity and memory footprint from quadratic to linear without changing its functionality: 
{\small
\begin{align}\label{eq:linearAtt}
    O_i &= \frac{\sum_{j=1}^N {\color{red}[\text{ReLU}(Q_i) \text{ReLU}(K_j)^T]} V_j}{\text{ReLU}(Q_i) \sum_{j=1}^N \text{ReLU}(K_j)^T} \nonumber \\ 
    &= \frac{\sum_{j=1}^N  {\color{blue}\text{ReLU}(Q_i)} {\color{red}[(\text{ReLU}(K_j)^T V_j)]}}{\text{ReLU}(Q_i) \sum_{j=1}^N \text{ReLU}(K_j)^T} \nonumber \\
    &= \frac{{\color{blue}\text{ReLU}(Q_i)} (\sum_{j=1}^N \text{ReLU}(K_j)^T V_j)}{\text{ReLU}(Q_i) (\sum_{j=1}^N \text{ReLU}(K_j)^T)}.
\end{align}
}As shown in Eq.~(\ref{eq:linearAtt}), we only need to compute $(\sum_{j=1}^N \text{ReLU}(K_j)^T V_j)$ $\in \mr^{d \times d}$ and $(\sum_{j=1}^N \text{ReLU}(K_j)^T)$ $\in \mr^{d \times 1}$ once, then can reuse them for each query, thereby only requires $\mathcal{O}(N)$ computational cost and $\mathcal{O}(N)$ memory.

Another key merit of ReLU linear attention is that it does not involve hardware-unfriendly operations like softmax, making it more efficient on hardware. For example, Figure~\ref{fig:latency_linear_vs_softmax} shows the latency comparison between softmax attention and ReLU linear attention. With similar computation, ReLU linear attention is significantly faster than softmax attention on the mobile CPU. 

\paragraph{Address ReLU Linear Attention's Limitations.} Although ReLU linear attention is superior to softmax attention in terms of computational complexity and hardware latency, ReLU linear attention has limitations. Figure~\ref{fig:linear_limitation} visualizes the attention maps of softmax attention and ReLU linear attention. Because of the lack of the non-linear similarity function, ReLU linear attention cannot generate concentrated attention maps, making it weak at capturing local information. 

To mitigate its limitation, we propose to enhance ReLU linear attention with convolution. Specifically, we insert a depthwise convolution in each FFN layer. An overview of the resulting building block is illustrated in Figure~\ref{fig:block} (left), where the ReLU linear attention captures context information and the FFN+DWConv captures local information. 

Furthermore, we propose to aggregate the information from nearby Q/K/V tokens to get multi-scale tokens to enhance the multi-scale learning ability of ReLU linear attention. This information aggregation process is independent for each Q, K, and V in each head. We only use small-kernel depthwise-separable convolutions \cite{howard2017mobilenets} for information aggregation to avoid hurting hardware efficiency. In the practical implementation, independently executing these aggregation operations is inefficient on GPU. Therefore, we take advantage of the group convolution to reduce the number of total operations. Specifically, all DWConvs are fused into a single DWConv while all 1x1 Convs are combined into a single 1x1 group convolution (Figure~\ref{fig:block} right) where the number of groups is $3\times \text{\#heads}$ and the number of channels in each group is d. After getting multi-scale tokens, we perform ReLU linear attention upon them to extract multi-scale global features. Finally, we concatenate the features along the head dimension and feed them to the final linear projection layer to fuse the features. 

\subsection{EfficientViT Architecture}
We build a new family of vision transformer models based on the proposed multi-scale linear attention module. The core building block (denoted as `EfficientViT Module') is illustrated in Figure~\ref{fig:block} (left). The macro architecture of EfficientViT is demonstrated in Figure~\ref{fig:macro_arch}. We use the standard backbone-head/encoder-decoder architecture design.
\begin{itemize}[leftmargin=*]
\item \textbf{Backbone.} The backbone of EfficientViT also follows the standard design, which consists of the input stem and four stages with gradually decreased feature map size and gradually increased channel number. We insert the EfficientViT module in Stages 3 and 4. For downsampling, we use an MBConv with stride 2. 

\item \textbf{Head.} P2, P3, and P4 denote the outputs of Stages 2, 3, and 4, forming a pyramid of feature maps. For simplicity and efficiency, we use 1x1 convolution and standard upsampling operation (e.g., bilinear/bicubic upsampling) to match their spatial and channel size and fuse them via addition. Since our backbone already has a strong context information extraction capacity, we adopt a simple head design that comprises several MBConv blocks and the output layers (i.e., prediction and upsample). In the experiments, we empirically find this simple head design is sufficient for achieving SOTA performances. 

In addition to dense prediction, our model can be applied to other vision tasks, such as image classification, by combining the backbone with task-specific heads. 
\end{itemize}

Following the same macro architecture, we design a series of models with different sizes to satisfy various efficiency constraints. We name these models EfficientViT-B0, EfficientViT-B1, EfficientViT-B2, and EfficientViT-B3, respectively. In addition, we designed the EfficientViT-L series for the cloud platforms. Detailed configurations of these models are provided in our official GitHub repository\footnote{\url{https://github.com/mit-han-lab/efficientvit}}.

%% file: sections/4-exp.tex
\section{Experiments}
\subsection{Setups}

\begin{table}[t]
\small\centering
\begin{tabular}{c c | c | c c }
\toprule
\multicolumn{2}{c|}{Components} & \multirow{2}{*}{mIoU $\uparrow$} & \multirow{2}{*}{Params $\downarrow$} & \multirow{2}{*}{MACs $\downarrow$} \\
\cmidrule{1-2}
Multi-scale & Global att. & & & \\
\midrule
& & 68.1 & 0.7M & 4.4G \\
\checkmark & & 72.3 & 0.7M & 4.4G \\
& \checkmark & 72.2 & 0.7M & 4.4G \\
\checkmark & \checkmark & \textbf{74.5} & 0.7M & 4.4G \\
\bottomrule
\end{tabular}
\caption{\textbf{Ablation Study.} The mIoU and MACs are measured on Cityscapes with 1024x2048 input resolution. We rescale the width of the models so that they have the same MACs. Multi-scale learning and the global receptive field are essential for obtaining good semantic segmentation performance. }\label{tab:ablation}
\end{table}

\begin{table*}[t]
\small\centering
\begin{tabular}{l | c c | c c | c c | c }
\toprule
\multirow{2}{*}{Models}  & \multirow{2}{*}{Top1 Acc $\uparrow$} & \multirow{2}{*}{Top5 Acc $\uparrow$} & \multirow{2}{*}{Params $\downarrow$} & \multirow{2}{*}{MACs $\downarrow$} & \multicolumn{2}{c|}{Latency $\downarrow$} & Throughput $\uparrow$ \\
& & & & & Nano(bs1) & Orin(bs1) & A100 (image/s) \\
\midrule
CoAtNet-0 \cite{dai2021coatnet} & 81.6 & - & 25M & 4.2G & 95.8ms & 4.5ms & 3011 \\
ConvNeXt-T \cite{liu2022convnet} & 82.1 & - & 29M & 4.5G & 87.9ms & 3.8ms & 3303 \\
\textbf{\modelc (r256)} & 82.7 & 96.1 & 24M & 2.1G & \textbf{58.5ms} & \textbf{2.8ms} & \textbf{5325} \\
\midrule
Swin-B \cite{liu2021swin} & 83.5 & - & 88M & 15G & 240ms & 6.0ms & 2236 \\
CoAtNet-1 \cite{dai2021coatnet} & 83.3 & - & 42M & 8.4G & 171ms & 8.3ms & 1512 \\
ConvNeXt-S \cite{liu2022convnet} & 83.1 & - & 50M & 8.7G & 146ms & 6.5ms & 2081 \\
\textbf{\modeld (r224)} & 83.5 & 96.4 & 49M & 4.0G & \textbf{101ms} & \textbf{4.4ms} & \textbf{3797} \\
\midrule
CoAtNet-2 \cite{dai2021coatnet} & 84.1 & - & 75M & 16G & 254ms & 10.3ms & 1174 \\
ConvNeXt-B \cite{liu2022convnet} & 83.8 & - & 89M & 15G & 211ms & 7.8ms & 1579 \\
\textbf{\modeld (r288)} & 84.2 & 96.7 & 49M & 6.5G & \textbf{141ms} & \textbf{5.6ms} & \textbf{2372} \\
\midrule
\midrule
CoAtNet-3 \cite{dai2021coatnet} & 84.5 & - & 168M & 35G & - & 15.4ms & 642 \\
ConvNeXt-L \cite{liu2022convnet} & 84.3 & - & 198M & 34G & - & 11.5ms & 1032 \\
EfficientNetV2-S \cite{tan2021efficientnetv2} & 83.9 & - & 22M & 8.8G & - & 4.3ms & 2869 \\
\textbf{\modelshort-L1 (r224)} & 84.5 & 96.9 & 53M & 5.3G & - & \textbf{2.6ms} & \textbf{6207} \\
\midrule
EfficientNetV2-M \cite{tan2021efficientnetv2} & 85.2 & - & 54M & 25G & - & 9.2ms & 1160 \\
FasterViT-4 \cite{hatamizadeh2023fastervit} & 85.4 & 97.3 & 425M & 37G & - & 13.0ms & 1382 \\
\textbf{\modelshort-L2 (r288)} & 85.6 & 97.4 & 64M & 11G & - & \textbf{4.3ms} & \textbf{3102} \\
\midrule
FasterViT-6 \cite{hatamizadeh2023fastervit} & 85.8 & 97.4 & 1360M & 142G & - & - & 594 \\
EfficientNetV2-L \cite{tan2021efficientnetv2} & 85.7 & - & 120M & 53G & - & - & 696 \\
\textbf{\modelshort-L2 (r384)} & 86.0 & 97.5 & 64M & 20G & - & - & \textbf{1784} \\
\bottomrule
\end{tabular}
\caption{\textbf{Backbone Performance on ImageNet Classification.} `r224' means the input resolution is 224x224. `bs1' represents that the latency is measured with batch size 1. }\label{tab:imagenet}
\end{table*}

\paragraph{Datasets.} We evaluate the effectiveness of EfficientViT on three representative high-resolution dense prediction tasks, including semantic segmentation, super-resolution, and Segment Anything.

For semantic segmentation, we use two popular benchmark datasets: Cityscapes \cite{cordts2016cityscapes} and ADE20K \cite{zhou2017scene}. In addition, we evaluate EfficientViT under two settings for super-resolution: lightweight super-resolution (SR) and high-resolution SR. We train models on DIV2K \cite{agustsson2017ntire} for lightweight SR and test on BSD100 \cite{martin2001database}. For high-resolution SR, we train models on the first 3000 training images of FFHQ \cite{karras2019style} and test on the first 500 validation images of FFHQ\footnote{\url{https://rb.gy/7je1a}}. 

Apart from dense prediction, we also study the effectiveness of EfficientViT for image classification using the ImageNet dataset \cite{deng2009imagenet}. 

\paragraph{Latency Measurement.} We measure the mobile latency on Qualcomm Snapdragon 8Gen1 CPU with Tensorflow-Lite\footnote{\url{https://www.tensorflow.org/lite}}, batch size 1 and fp32. We use TensorRT\footnote{\url{https://docs.nvidia.com/deeplearning/tensorrt/}} and fp16 to measure the latency on edge GPU and cloud GPU. The data transfer time is included in the reported latency/throughput results. 

\paragraph{Implementation Details.} We implement our models using Pytorch \cite{paszke2019pytorch} and train them on GPUs. We use the AdamW optimizer with cosine learning rate decay for training our models. For multi-scale linear attention, we use a two-branch design for the best trade-off between performance and efficiency, where 5x5 nearby tokens are aggregated to generate multi-scale tokens. 

For semantic segmentation experiments, we use the mean Intersection over Union (mIoU) as our evaluation metric. The backbone is initialized with weights pretrained on ImageNet and the head is initialized randomly, following the common practice. 

For super-resolution, we use PSNR and SSIM on the Y channel as the evaluation metrics, same as previous work \cite{liang2021swinir}. The models are trained with random initialization.

\subsection{Ablation Study}
\paragraph{Effectiveness of EfficientViT Module.}

We conduct ablation study experiments on Cityscapes to study the effectiveness of two key design components of our EfficientViT module, i.e., multi-scale learning and global attention. To eliminate the impact of pre-training, we train all models from random initialization. In addition, we rescale the width of the models so that they have the same \#MACs. The results are summarized in Table~\ref{tab:ablation}. We can see that removing either global attention or multi-scale learning will significantly hurt the performances. It shows that all of them are essential for achieving a better trade-off between performance and efficiency. 

\paragraph{Backbone Performance on ImageNet.}
To understand the effectiveness of EfficientViT's backbone in image classification, we train our models on ImageNet following the standard training strategy. We summarize the results and compare our models with SOTA image classification models in Table~\ref{tab:imagenet}. 

Though EfficientViT is designed for high-resolution dense prediction, it achieves highly competitive performances on ImageNet classification. In particular, EfficientViT-L2-r384 obtains 86.0 top1 accuracy on ImageNet, providing +0.3 accuracy gain over EfficientNetV2-L and 2.6x speedup on A100 GPU. 

\begin{table*}[t]
\small\centering
\begin{tabular}{l | c | c c | c c | c }
\toprule
\multirow{2}{*}{Models} & \multirow{2}{*}{mIoU $\uparrow$} & \multirow{2}{*}{Params $\downarrow$} & \multirow{2}{*}{MACs $\downarrow$} & \multicolumn{2}{c|}{Latency $\downarrow$} & Throughput $\uparrow$ \\
& & & & Nano(bs1) & Orin(bs1) & A100(image/s) \\
\midrule
DeepLabV3plus-Mbv2 \cite{chen2018encoder} & 75.2 & 15M & 555G & - & 83.5ms & 102 \\
\textbf{\modela} & 75.7 & 0.7M & 4.4G & \textbf{0.28s} & \textbf{9.9ms} & \textbf{263} \\
\midrule
SegFormer-B1 \cite{xie2021segformer} & 78.5 & 14M & 244G & 5.6s & 146ms & 49 \\
SegNeXt-T \cite{guo2022segnext} & 79.8 & 4.3M & 51G & 2.2s & 93.2ms & 95 \\
\textbf{\modelb} & 80.5 & 4.8M & 25G & \textbf{0.82s} & \textbf{24.3ms} & \textbf{175} \\
\midrule
SegFormer-B3 \cite{xie2021segformer} & 81.7 & 47M & 963G & - & 407ms & 18 \\
SegNeXt-S \cite{guo2022segnext} & 81.3 & 14M & 125G & 3.4s & 127ms & 70 \\
\textbf{\modelc} & 82.1 & 15M & 74G & \textbf{1.7s} & \textbf{46.5ms} & \textbf{112} \\
\midrule
SegFormer-B5 \cite{xie2021segformer} & 82.4 & 85M & 1460G & - & 638ms & 12 \\
SegNeXt-B \cite{guo2022segnext} & 82.6 & 28M & 276G & - & 228ms & 41 \\
\textbf{\modeld} & 83.0 & 40M & 179G & - & 81.8ms & 70 \\
\textbf{\modelshort-L1} & 82.7 & 40M & 282G & - & \textbf{45.9ms} & \textbf{122} \\
\midrule
SegNeXt-L \cite{guo2022segnext} & 83.2 & 49M & 578G & - & 374ms & 26 \\
\textbf{\modelshort-L2} & 83.2 & 53M & 396G & - & \textbf{60.0ms} & \textbf{102} \\
\bottomrule
\end{tabular}
\caption{\textbf{Comparison with SOTA Semantic Segmentation Models on Cityscapes.} The input resolution is 1024x2048 for all models. Models with similar mIoU are grouped for efficiency comparison. }\label{tab:cityscapes_main}
\end{table*}

\begin{table*}[t]
\small\centering
\begin{tabular}{l | c | c c | c c | c }
\toprule
\multirow{2}{*}{Models} & \multirow{2}{*}{mIoU $\uparrow$} & \multirow{2}{*}{Params $\downarrow$} & \multirow{2}{*}{MACs $\downarrow$} & \multicolumn{2}{c|}{Latency $\downarrow$} & Throughput $\uparrow$ \\
& & & & Nano(bs1) & Orin(bs1) & A100(image/s) \\
\midrule
SegFormer-B1 \cite{xie2021segformer} & 42.2 & 14M & 16G & 389ms & 12.3ms & 542 \\
SegNeXt-T \cite{guo2022segnext} & 41.1 & 4.3M & 6.6G & 281ms & 12.4ms & 842 \\
\textbf{\modelb} & 42.8 & 4.8M & 3.1G & \textbf{110ms} & \textbf{4.0ms} & \textbf{1142} \\
\midrule
SegNeXt-S \cite{guo2022segnext} & 44.3 & 14M & 16G & 428ms & 17.2ms & 592 \\
\textbf{\modelc} & 45.9 & 15M & 9.1G & \textbf{212ms} & \textbf{7.3ms} & \textbf{846} \\
\midrule
Mask2Former \cite{cheng2022masked} & 47.7 & 47M & 74G & - & - & -\\
MaskFormer \cite{cheng2021per} & 46.7 & 42M & 55G & - & - & - \\
SegFormer-B2 \cite{xie2021segformer} & 46.5 & 28M & 62G & 920ms & 24.3ms & 345 \\
SegNeXt-B \cite{guo2022segnext} & 48.5 & 28M & 35G & 806ms & 32.9ms & 347 \\
\textbf{\modeld} & 49.0 & 39M & 22G & 411ms & 12.5ms & 555 \\
\textbf{\modelshort-L1} & 49.2 & 40M & 36G & - & \textbf{7.2ms} & \textbf{947} \\
\midrule
SegFormer-B4 \cite{xie2021segformer} & 50.3 & 64M & 96G & - & 44.9ms & 212 \\
\textbf{\modelshort-L2} & 50.7 & 51M & 45G & - & \textbf{9.0ms} & \textbf{758} \\
\bottomrule
\end{tabular}
\caption{\textbf{Comparison with SOTA Semantic Segmentation Models on ADE20K.} The shorter side of the image is resized to 512, following the common practice.}\label{tab:ade20k_main}
\vspace{-10pt}
\end{table*}

\begin{table*}[t]
\small\centering
\begin{tabular}{l | c c | c c | c c | c c }
\toprule
\multirow{2}{*}{Model} & \multicolumn{4}{c|}{FFHQ (512x512 $\rightarrow$ 1024x1024)} & \multicolumn{4}{c}{BSD100 (160x240 $\rightarrow$ 320x480)} \\
\cmidrule{2-9}
 & PSNR $\uparrow$ & SSIM $\uparrow$ & A100(bs1) $\downarrow$ & Speedup $\uparrow$ & PSNR $\uparrow$ & SSIM $\uparrow$ & A100(bs1) $\downarrow$ & Speedup $\uparrow$ \\
\midrule
Restormer \cite{zamir2022restormer} & 43.43 & 0.9806 & 92.0ms & 1x & 32.31 & \textbf{0.9021} & 15.1ms & 1x \\
SwinIR \cite{liang2021swinir} & 43.49 & 0.9807 & 61.2ms & 1.5x & 32.31 & 0.9012 & 9.7ms & 1.6x \\
VapSR \cite{zhou2022efficient} & - & - & - & - & 32.27 & 0.9011 & 4.8ms & 3.1x \\
BSRN \cite{li2022blueprint} & - & - & - & - & 32.24 & 0.9006 & 4.5ms & 3.4x \\
\midrule
\textbf{\modelshort w0.75} & 43.54 & 0.9809 & \textbf{14.3ms} & \textbf{6.4x} & 32.31 & 0.9016 & \textbf{2.8ms} & \textbf{5.4x} \\
\textbf{\modelshort} & \textbf{43.58} & \textbf{0.9810} & 17.8ms & 5.2x & \textbf{32.33} & 0.9019 & 3.2ms & 4.7x \\
\bottomrule
\end{tabular}
\caption{\textbf{Comparison with SOTA super-resolution models.} }\label{tab:super_res_main}
\end{table*}

\subsection{Semantic Segmentation} 

\paragraph{Cityscapes.} Table~\ref{tab:cityscapes_main} reports the comparison between EfficientViT and SOTA semantic segmentation models on Cityscapes. EfficientViT achieves remarkable efficiency improvements over prior SOTA semantic segmentation models without sacrificing performances. Specifically, compared with SegFormer, EfficientViT obtains up to 13x \#MACs saving and up to 8.8x latency reduction on the edge GPU (Jetson AGX Orin) with higher mIoU. 
Compared with SegNeXt, EfficientViT provides up to 2.0x MACs reduction and 3.8x speedup on the edge GPU (Jetson AGX Orin) while maintaining higher mIoU. On A100 GPU, EfficientViT delivers up to 3.9x higher throughput than SegNeXt and 10.2x higher throughput than SegFormer while achieving the same or higher mIoU. Having similar computational cost, EfficientViT also yields significant performance gains over previous SOTA models. For example, EfficientViT-B3 delivers +4.5 mIoU gain over SegFormer-B1 with lower MACs. 

\begin{figure}[t]
    \centering
    \includegraphics[width=\linewidth]
    {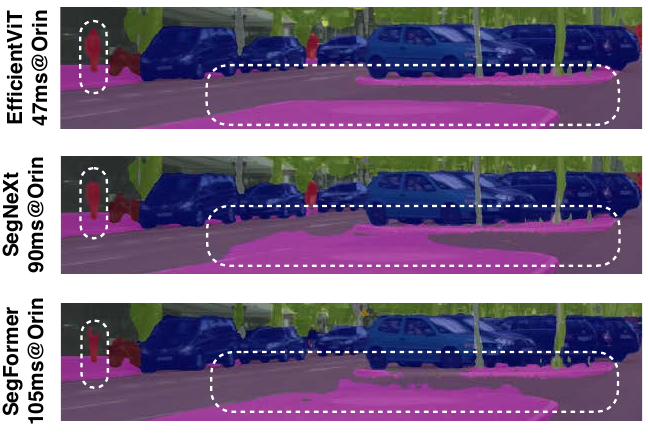}
    \caption{\textbf{Qualitative results on Cityscapes.}}
    \label{fig:visualization}
\end{figure}

In addition to the quantitative results, we visualize EfficientViT and the baseline models qualitatively on Cityscapes. The results are shown in Figure~\ref{fig:visualization}. We can find that EfficientViT can better recognize boundaries and small objects than the baseline models while achieving lower latency on GPU. 

\vspace{-10pt}
\paragraph{ADE20K.} Table~\ref{tab:ade20k_main} summarizes the comparison between EfficientViT and SOTA semantic segmentation models on ADE20K. Like Cityscapes, we can see that EfficientViT also achieves significant efficiency improvements on ADE20K. For example, with +0.6 mIoU gain, EfficientViT-B1 provides 5.2x MACs reduction and up to 3.5x GPU latency reduction than SegFormer-B1. With +1.6 mIoU gain, EfficientViT-B2 requires 1.8x fewer computational costs and runs 2.4x faster on Jetson AGX Orin GPU than SegNeXt-S. 

\subsection{Super-Resolution}
Table~\ref{tab:super_res_main} presents the comparison of \modelshort with SOTA ViT-based SR methods (SwinIR \cite{liang2021swinir} and Restormer \cite{zamir2022restormer}) and SOTA CNN-based SR methods (VapSR \cite{zhou2022efficient} and BSRN \cite{li2022blueprint}). \modelshort provides a better latency-performance trade-off than all compared methods. 

On lightweight SR, \modelshort provides up to 0.09dB gain in PSNR on BSD100 while maintaining the same or lower GPU latency compared with SOTA CNN-based SR methods. Compared with SOTA ViT-based SR methods, \modelshort provides up to 5.4$\times$ speedup on GPU and maintains the same PSNR on BSD100. 

\begin{figure}[t]
    \centering
    \includegraphics[width=\linewidth]
    {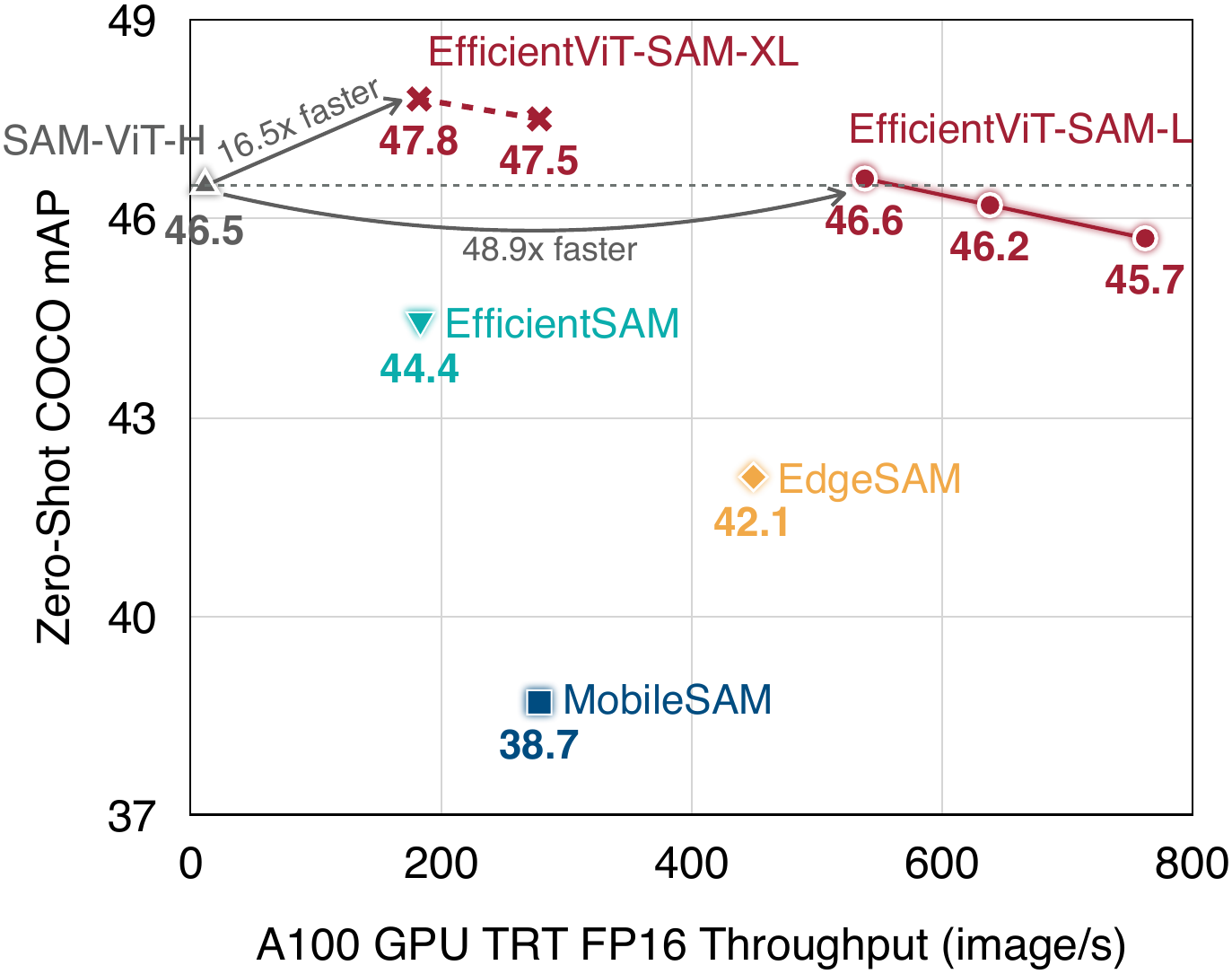}
    \caption{\textbf{Throughput vs. COCO Zero-Shot Instance Segmentation mAP.} EfficientViT-SAM is the first accelerated SAM model that matches/outperforms SAM-ViT-H's \cite{kirillov2023segment} zero-shot performance, delivering the SOTA performance-efficiency trade-off.}
    \label{fig:sam_curve}
\end{figure}

On high-resolution SR, the advantage of \modelshort over previous ViT-based SR methods becomes more significant. Compared with Restormer, \modelshort achieves up to 6.4$\times$ speedup on GPU and provides 0.11dB gain in PSNR on FFHQ. 

\begin{table*}[t]
\small\centering
\setlength{\tabcolsep}{5pt}
\begin{tabular}{l | c c | c | c c c c | c c c c}
\toprule
& \multirow{2}{*}{Params $\downarrow$} & \multirow{2}{*}{MACs $\downarrow$} &  Throughput $\uparrow$  & \multicolumn{4}{c|}{COCO} & \multicolumn{4}{c}{LVIS}  \\
\cmidrule{5-12}
& & & A100(image/s) & $\rm mAP$ & $\rm AP^S$ & $\rm AP^M$ & $\rm AP^L$ & $\rm mAP$ & $\rm AP^S$ & $\rm AP^M$ & $\rm AP^L$ \\
\midrule
SAM-ViT-H \cite{kirillov2023segment} & 641M & 2973G & 11 & 46.5 & 30.8 & 51.0 & 61.7 & 44.2 & 31.8 & 57.1 & 65.3  \\
\midrule
EfficientViT-SAM-L0 & 35M & 35G & 762 & 45.7 & 28.2 & 49.5 & 63.4 & 41.8 & 28.8 & 53.4 & 64.7 \\
EfficientViT-SAM-L1 & 48M & 49G & 638 & 46.2 & 28.7 & 50.4 & 64.0 & 42.1 & 29.1 & 54.3 & 65.0\\
EfficientViT-SAM-L2 & 61M & 69G & 538 & 46.6 & 28.9 & 50.8 & 64.2 & 42.7 & 29.4 & 55.1 & 65.5 \\
\midrule 
EfficientViT-SAM-XL0 & 117M & 185G & 278 & 47.5 & 30.0 & 51.5 & 64.6 & 43.9 & 31.2 & 56.2 & 65.9\\
EfficientViT-SAM-XL1 & 203M & 322G & 182 & 47.8 & 30.5 & 51.8 & 64.7 & 44.4 & 31.6 & 57.0 & 66.4\\
\bottomrule
\end{tabular}
\caption{\textbf{Zero-Shot Instance Segmentation Results, Prompted with ViTDet Boxes.} Throughput is profiled on A100 GPU with TensorRT and fp16, including the image encoder and SAM head.}
\vspace{-10pt}
\label{table:sam_box_vitdet}
\end{table*}

\begin{table}[t]
\small\centering
\setlength{\tabcolsep}{2pt}
\scalebox{0.85}{
\begin{tabular}{l|ccc|ccc}
\toprule
 & \multicolumn{3}{c|}{COCO} & \multicolumn{3}{c}{LVIS}  \\
\midrule
 & 1 click & 3 click & 5 click & 1 click & 3 click & 5 click\\
\midrule
SAM-ViT-H \cite{kirillov2023segment} & 58.4  & 69.6  & 71.4  & 59.2  & 66.0 & 66.8  \\
\midrule 
EfficientViT-SAM-XL1 & 59.8 & 71.3 & 75.3 & 56.6 & 67.0 & 71.7\\
\bottomrule
\end{tabular}}
\caption{\textbf{Zero-Shot Point-Prompted Segmentation Results.}}
\label{table:sam_point}
\vspace{-10pt}
\end{table}

\subsection{Segment Anything}
We build EfficientViT-SAM, a new family of accelerated segment anything models, by leveraging EfficientViT to replace SAM's image encoder. Meanwhile, we retain SAM's lightweight prompt encoder and mask decoder. The training process consists of two phases. First, we train the image encoder of EfficientViT-SAM using SAM's image encoder as the teacher. Second, we train EfficientViT-SAM end-to-end using the whole SA-1B dataset \cite{kirillov2023segment}.

We thoroughly test EfficientViT-SAM on various zero-shot benchmarks to verify its effectiveness. Table~\ref{table:sam_box_vitdet} demonstrates the zero-shot instance segmentation results on COCO \cite{lin2014microsoft} and LVIS \cite{gupta2019lvis}, prompted with the predicted bounding boxes from ViTDet \cite{li2022exploring}. EfficientViT-SAM provides superior performance/efficiency compared with SAM-ViT-H \cite{kirillov2023segment}. In particular, EfficientViT-SAM-XL1 outperforms SAM-ViT-H on COCO and LVIS while having 16.5$\times$ higher throughput on A100 GPU. 

Figure~\ref{fig:sam_curve} shows the comparison between EfficientViT-SAM and prior SAM models. EfficientViT-SAM is the first accelerated SAM model that matches/outperforms SAM-ViT-H's \cite{kirillov2023segment} zero-shot performance, delivering the SOTA performance-efficiency trade-off.

Apart from box-prompted instance segmentation, we also evaluate EfficientViT-SAM on point-prompted segmentation. The results are summarized in Table~\ref{table:sam_point}. EfficientViT-SAM-XL1 outperforms SAM-ViT-H in most cases, especially when more points are given. On LVIS, when given a single point, we find SAM-ViT-H performs better than EfficientViT-SAM-XL1. This might be because we do not have the interactive segmentation setup during the end-to-end training phase. Further investigation is needed to improve the performance of the single-point setting.

%% file: sections/2-related-work.tex
\section{Related Work}

\paragraph{High-Resolution Dense Prediction.} 

Dense prediction targets producing predictions for each pixel given the input image. It can be viewed as an extension of image classification from per-image prediction to per-pixel predictions. Extensive studies have been done to improve the performance of CNN-based high-resolution dense prediction models \cite{badrinarayanan2017segnet,ronneberger2015u,chen2017deeplab,zhao2017pyramid,yuan2020object,wang2020deep}. 

In addition, there are also some works targeting improving the efficiency of high-resolution dense prediction models \cite{zhao2018icnet,poudel2019fast,li2019dfanet,yu2018bisenet}. While these models provide good efficiency, their performances are far behind SOTA high-resolution dense prediction models. 

Compared to these works, our models provide a better trade-off between performance and efficiency by enabling a global receptive field and multi-scale learning with lightweight operations. 

\paragraph{Efficient Vision Transformer.} 
While ViT provides impressive performances in the high-computation region, it is usually inferior to previous efficient CNNs \cite{tan2019efficientnet,howard2019searching,cai2020once,han2020ghostnet} when targeting the low-computation region. To close the gap, MobileViT \cite{mehta2022mobilevit} proposes to combine the strength of CNN and ViT by replacing local processing in convolutions with global processing using transformers. MobileFormer \cite{chen2021mobile} proposes to parallelize MobileNet and Transformer with a two-way bridge in between for feature fusing. NASViT \cite{gong2022nasvit} proposes to leverage neural architecture search to search for efficient ViT architectures. 

However, these models mainly focus on image classification and still rely on softmax attention with quadratic computational complexity, thus unsuitable for high-resolution dense prediction. 

\paragraph{Efficient Deep Learning.} Our work is also related to efficient deep learning, which aims at improving the efficiency of deep neural networks so that we can deploy them on hardware platforms with limited resources, such as mobile phones and IoT devices. Typical technologies in efficient deep learning include network pruning \cite{han2015learning,he2017channel,liu2017learning}, quantization \cite{han2016deep}, efficient model architecture design \cite{howard2017mobilenets,ma2018shufflenet}, and training techniques \cite{hinton2015distilling,cai2021network,cai2020tinytl}. In addition to manual designs, many recent works use AutoML techniques \cite{zoph2017neural,cai2018efficient,cai2019proxylessnas} to automatically design \cite{cai2020once}, prune \cite{he2018amc} and quantize \cite{wang2020apq} neural networks.

%% file: sections/5-conclusion.tex
\section{Conclusion}
In this work, we studied efficient architecture design for high-resolution dense prediction. We introduced a lightweight multi-scale attention module that simultaneously achieves a global receptive field, and multi-scale learning with lightweight and hardware-efficient operations, thus providing significant speedup on diverse hardware devices without performance loss than SOTA high-resolution dense prediction models. For future work, we will explore applying EfficientViT to other vision tasks and further scaling up our EfficientViT models. 

\section*{Acknowledgments}
We thank MIT-IBM Watson AI Lab, MIT AI Hardware Program, Amazon and MIT Science Hub, Qualcomm Innovation Fellowship, National Science Foundation for supporting this research. 